\documentclass[letterpaper, 10 pt, conference]{ieeeconf}  
\usepackage{amsmath,amsfonts,amssymb,mathrsfs}
\usepackage{algorithmic}
\usepackage{algorithm}
\usepackage{array}
\usepackage[caption=false,font=normalsize,labelfont=sf,textfont=sf]{subfig}
\usepackage{textcomp}
\usepackage{stfloats}
\usepackage{url}
\usepackage{verbatim}
\usepackage{graphicx}
\usepackage{cite}
\usepackage{supertabular}
\usepackage[table,xcdraw]{xcolor}
\usepackage{lipsum}
\usepackage{makecell}
\usepackage{pifont}
\usepackage{orcidlink}
\usepackage{hyperref}
\hypersetup{colorlinks=true,
	linkcolor=blue,
	anchorcolor=blue,
	citecolor=blue}
\newcommand{\mb}[1]{\mathbf{#1}} 
\newcommand{\bs}[1]{\boldsymbol{#1}}
\newcommand{\mr}[1]{\mathrm{#1}}

\newcommand{\mrd}{\mathrm{d}}

\IEEEoverridecommandlockouts                              

\title{\LARGE \bf
Topology-Inspired Morphological Descriptor for Soft Continuum Robots
}

\author{
Zhiwei Wu$^{\orcidlink{https://orcid.org/0000-0002-3957-3063}}$,~\IEEEmembership{Graduate Student Member,~IEEE}, Jiahao~Luo$^{\orcidlink{0009-0003-3197-1782}}$, Siyi~Wei$^{\orcidlink{0009-0005-9967-4025}}$, and Jinhui~Zhang$^{\orcidlink{0000-0002-2405-894X}}$.
\thanks{
Z. Wu, S. Wei, J. Luo, and J. Zhang are with the School of Automation, Beijing Institute of Technology, Beijing 10081, China. (e-mails: zhiweiwu.cn@outlook.com, siyi.wei@bit.edu.cn, lawkaho.edu@outlook.com).
}
\thanks{*Corresponding author: Jinhui Zhang (e-mail: zhangjinh@bit.edu.cn).}
}

\begin{document}

\maketitle

\thispagestyle{empty}
\pagestyle{empty}

\begin{abstract}

This paper presents a topology-inspired morphological descriptor for soft continuum robots by combining a pseudo-rigid-body (PRB) model with Morse theory to achieve a quantitative characterization of robot morphologies. By counting critical points of directional projections, the proposed descriptor enables a discrete representation of multimodal configurations and facilitates morphological classification. Furthermore, we apply the descriptor to morphology control by formulating the target configuration as an optimization problem to compute actuation parameters that generate equilibrium shapes with desired topological features. The proposed framework provides a unified methodology for quantitative morphology description, classification, and control of soft continuum robots, with the potential to enhance their precision and adaptability in medical applications such as minimally invasive surgery and endovascular interventions.


\end{abstract}

\section{INTRODUCTION}


The ability to achieve diverse and controllable morphologies is a key advantage of continuum robots, particularly in medical applications such as endovascular intervention, minimally invasive surgery, and navigation within confined anatomical pathways\cite{burgner2015continuum, shi2016shape}. These tasks require robots to adapt their shape dynamically to traverse tortuous lumens, avoid sensitive tissues, and accurately reach target sites. However, existing approaches to describing multimodal configurations are largely qualitative, using intuitive labels such as “J-shape” or “S-shape” \cite{wei2025multimodal,linMagneticContinuumRobot2021,huang2024design,wang2023magnetic,cao2023magnetic} Such descriptions lack a rigorous quantitative basis and do not reveal the intrinsic geometric–topological factors that govern morphology generation. Consequently, it is difficult to establish systematic control strategies that can realize desired shapes in a predictable manner. To address these challenges, we propose a topology-inspired morphological descriptor that quantitatively characterizes continuum robot configurations. We first introduce a pseudo-rigid-body (PRB) model to provide an efficient kinematic representation, and then leverage Morse theory to extract topological invariants of robot shapes. Based on this framework, we establish a descriptor that can support both objective morphology classification and control-oriented applications.


\section{Kinematics: Geometric Representation}

\begin{figure}[t]
    \centering
    \includegraphics[width=\columnwidth]{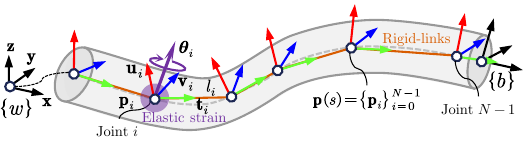}
    \caption{Illustration of the pseudo-rigid-body (PRB) model describing the SCR. 
    }
    \label{fig:1}
\end{figure}

Morphological analysis of soft continuum robots (SCRs) is generally performed at their equilibrium configurations. We first introduce a geometric representation that characterizes such equilibrium states.
The SCR is modeled using the efficient PRB model \cite{howell2001compliant,roesthuisSteeringMultisegmentContinuum2016,wu2025unified}. Let the centerline of the SCR be given by a smooth embedding $\mb{p}(s):[0,L]\to\mathbb{R}^3$, where $s$ is the arc-length parameter, and $L$ is the length of the SCR with inextensible assumption. As shown in Fig.~\ref{fig:1}, the centerline is descritized to $N$ points of interest $\mb{p}_i\in\mathbb{R}^3$, connected rigid links with length $l_i,i=0,\dots,N-1$. Each joint is attached with an orthogonal local frame $\{\bar{\mb{t}}_i,\bar{\mb{u}}_i,\bar{\mb{v}}_i=\bar{\mb{t}}_i\times\bar{\mb{u}}_i\}$, where the overbars indicate a straightended reference state, and the tangent vector $\bar{\mb{t}}_i$ remains aligned with the successive link. The rotation axis $\bs\theta_i\in\mathbb{R}^3$ is determined through the local frame. The associated rotation matrix is defined as $\mb{R}_i(\bs\theta_i)=\mb{e}^{[\bs\theta_i]_\times}\in \mr{SO}(3)$, and we define a useful shorthand notation $(\mb{R}_i^j)$ to represent the product of rotation matrices that $\mb{R}_i^j=\mb{R}_i\mb{R}_{i+1}\cdots\mb{R}_j$. Local frame axis in deformed states, e.g., the tangent vector, can be easily computed by $\mb{t}_i=\mb{R}_0^i\bar{\mb{t}}_i$. 


We denote the configuration space of the SCR as $\bs\theta=[\bs\theta_0,\dots,\bs\theta_{N-1}]^\top\in\mathbb{R}^{3N}$.
Let $\{w\}$ be the world frame attached to joint $0$ and frame $\{b\}$ to the distal end. The associated distal pose of SCR in reference state is denoted as $_{w}^{b}\bar{\mb{H}}\in\mr{SE}(3)$. 
The static equilibrium equation can be derived from the Euler-Lagrange formulation by considering the potential energy of the joints and the virtual work of external forces. The configuration at equilibrium minimizes the total potential energy, which leads to the force-balance relationship \(\bs{\Lambda}(\bs{\theta}-\bar{\bs\theta})+\bs\tau_{int}(\bs\theta)=\bs\tau_{ext}(\bs\theta)\), where $\bs{\Lambda}$ denotes the diagonal stiffness matrix, $\bs{\tau}_{int}(\bs\theta)$ represents the nonlinear internal torques, and $\bs{\tau}_{ext}(\bs\theta)$ denotes the active and external torques, including those generated by actuators (e.g., tendons\cite{kato2014tendon}, hydraulic pressure\cite{ikuta2006multi}, magnetic fields\cite{linMagneticContinuumRobot2021,kimTeleroboticNeurovascularInterventions2022}) and external loads.

\section{Morphological Descriptor}

We now turn our attention to the morphological descriptor. We begin by recalling some basic concepts of Morse theory. 
Let $f:\mathbb{R}\to\mathbb{R}$ be a Morse function on the one-dimensional manifold. The central idea of Morse theory is to extract topological invariants of the underlying space by counting critical points of such functions \cite{milnor1963morse}. A canonical and physically interpretable choice of $f$ is the directional projection: $f = \mb{p}(s)^\top\mathbf{v}$, where $\mathbf{v} \in \mathbb{S}^2$ is a fixed unit vector. A classical result from Morse theory asserts that such projection functions are Morse for almost all choices of $\mb{v}$ \cite{audin2014morse}. 
The morphological descriptor can naturally be defined by counting the non-degenerate critical points of $f$:
\begin{equation*}
\mathfrak{M}_{\mb{v}}(\mb{p}) \triangleq \# \left\{ s \in (0, L)\ \middle|\ \mathbf{v}^\top\frac{\mrd\mb{p}}{\mrd s} = 0,\ \mathbf{v}^\top\frac{\mrd^2\mb{p}}{\mrd s^2} \neq 0 \right\},
\end{equation*}
where $\frac{\mrd\mb{p}}{\mrd s}$ and $\frac{\mrd^2\mb{p}}{\mrd s^2}$ are also referred to as the tangent vector $\mb{t}(s)$ and curvature vector $\bs{\kappa}(s)$, respectively.
Recall that the PRB model represents the curve $\mb{p}(s)$ by a sequence of points $\{\mb{p}_i\}_{i=0}^{N-1}$ sampled at uniform arc-length intervals. 
The discrete description of these vectors are $\mb{t}_i=\mb{R}_0^i\bar{\mb{t}}_i$ and \(\bs{\kappa}_i = (\mb{t}_i-\mb{t}_{i-1})/l_i=(\mb{R}_{0}^i\bar{\mb{t}}_i-\mb{R}_0^{i-1}\bar{\mb{t}}_{i-1})/l_i\approx[\bs\theta_i]_\times\bar{\mb{t}}_i/l_i\), 
yielding the discrete form of the morphological descriptor:
\begin{equation*}
    \mathfrak{M}_{\mb{v}}(\bs\theta)\triangleq\#\{i\in(0,N)|\mb{v}^\top\mb{R}_0^i(\bs\theta)\bar{\mb{t}}_i=0, \mb{v}^\top[\bs\theta_i]_\times\bar{\mb{t}}_i\neq 0\}.
\end{equation*}
In practice, the criterion $\mb{v}^\top\mb{R}_0^i(\bs\theta)\bar{\mb{t}}_i=0$ will be modified as $(\mb{v}^\top\mb{t}_{i})(\mb{v}^\top\mb{t}_{i+1})<0$ to avoid numerical issues. Under our definition, shapes with the same number of critical points $\mathfrak{M}_\mathbf{v}(\boldsymbol\theta)$, known as the Morse number, are considered morphologically equivalent. For centerlines modeled as compact curves on $[0, L]$, Morse theory guarantees that non-degenerate critical points are isolated and thus finite \cite{milnor1963morse}. This enables discrete and comparable morphological classification. Note that morphological equivalence does not imply topological equivalence: any continuous, self-avoiding open curve of finite length is topologically homeomorphic to $[0,1]$ \cite{edelsbrunner2010computational}. Another worth noting is that the Morse number may vary depending on the directional projection. For planar curves, the projector can be naturally chosen, whereas the spatial curves have no canonical projection direction. Therefore, one can choose $\mb{v}^*=\arg\max_{\mb{v}\in\mathbb{S}^2} \mathfrak{M}_{\mb{v}}(\bs\theta)$, direction of initial state, direction that orthogonal to the distal end, or other direction of interest.

\section{Applications}

\subsection{Morphological Classification}

\begin{figure}
    \centering
    \includegraphics[width=\columnwidth]{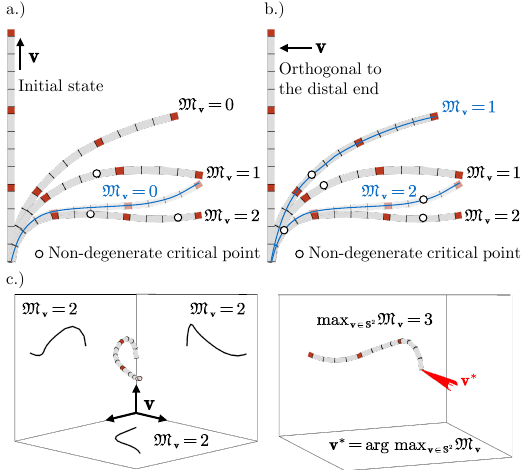}
    \caption{Morphological classification using proposed descriptor.}
    \label{fig:2}
\end{figure}

Here we present morphological classification for SCRs based on the proposed descriptor. The SCR is asummed to be natural straight. As shown in Fig.~\ref{fig:2}, the SCR contains three embedded actuation elements (marked in red). The Morse number with projection direction of initial state (Fig.~\ref{fig:2}a), orthogonal to the distal end (Fig.~\ref{fig:2}a), of maximized descriptor (Fig.~\ref{fig:2}a) are compared. Fig.~\ref{fig:2}a shows a classification result similar to related work \cite{wei2025multimodal,cao2023magnetic,wang2023magnetic}. Morse numbers 0, 1, and 2 correspond to $J$-, $C$-, and $S$-shaped configurations, respectively, and the proposed descriptor resolves ambiguous cases. Morse theory indicates that a critical point count of 0 implies a shape homeomorphic to the initial state, i.e., with small deflection. A counterintuitive case, highlighted in blue, is classified as $J$-shaped, meaning it is regarded as morphologically equivalent to a $C$-shape under this projection. Fig.~\ref{fig:2}b shows a projection direction that avoids this counterintuitive result: $J$- and $C$-shapes are then considered equivalent, while $S$-shapes are classified as expected. Fig.~\ref{fig:2}c illustrates that a cluster of projection directions may exist that maximizes the Morse number, obtainable via spherical search. Notably, the world coordinate axes, though straightforward choices, may not lie in this cluster.





\subsection{Morphology Control}

A potential application of the proposed morphological descriptor is in morphology control, which can be formulated as the inverse of the morphological classification problem. 
We aim to determine the actuation parameters $\mb{u}$ that generate external torques $\bs\tau(\bs\theta,\mb{u})$ such that a specified joint $i$ becomes a non-degenerate critical point of the descriptor.
This formulation enables the robot to attain a desired equilibrium morphology that exhibits the targeted topological feature. The optimisation problem is defined as
\begin{equation}
\label{eqn:inverseproblem}
    \begin{aligned}
        \min _\mb{u}&\ J(u)=\left(\mathbf{v}^{\top} \mathbf{R}_0^i\left(\bs{\theta}\right) \overline{\mathbf{t}}_i\right)^2+\frac{\alpha}{\left(\mathbf{v}^{\top}\left[\bs{\theta}\right]_{\times} \overline{\mathbf{t}}_i\right)^2+\epsilon}\\
        \mathrm{s.t.}&\ \bs{\Lambda}(\bs{\theta}-\bar{\bs\theta})+\bs\tau_{int}(\bs\theta)=\bs\tau_{ext}(\bs\theta,\mb{u})
    \end{aligned}
\end{equation}
where the first term enforces the critical-point condition by driving
$\mathbf{v}^{\top} \mathbf{R}_0^i(\boldsymbol{\theta}) \overline{\mathbf{t}}_i$ to zero, and the second term penalises degeneracy using a reciprocal barrier weighted by $\alpha$ with a small $\epsilon>0$ to avoid numerical singularities.
The equality constraint is the static equilibrium equation, which couples the configuration $\boldsymbol{\theta}$ with the actuation parameters through the internal and external torques.
The actuation command $\mathbf{u}^\ast$ can be obtaind by solving \eqref{eqn:inverseproblem} ro realise the desired non-degenerate critical point, thereby shaping the SCR into the target equilibrium morphology.


\section{Conclusion}
This paper presented a topology-inspired morphological descriptor for soft continuum robots, future work may extend this framework to dynamic scenarios, incorporate real-time sensing feedback, and explore broader applications in multi-segment or hybrid soft–rigid robotic systems.

\addtolength{\textheight}{-12cm}   







\newpage
\bibliographystyle{ieeetr}
\bibliography{references}

\begin{thebibliography}{10}

\bibitem{burgner2015continuum}
J.~Burgner-Kahrs, D.~C. Rucker, and H.~Choset, ``Continuum robots for medical
  applications: A survey,'' {\em IEEE transactions on robotics}, vol.~31,
  no.~6, pp.~1261--1280, 2015.

\bibitem{shi2016shape}
C.~Shi, X.~Luo, P.~Qi, T.~Li, S.~Song, Z.~Najdovski, T.~Fukuda, and H.~Ren,
  ``Shape sensing techniques for continuum robots in minimally invasive
  surgery: A survey,'' {\em IEEE Transactions on Biomedical Engineering},
  vol.~64, no.~8, pp.~1665--1678, 2016.

\bibitem{wei2025multimodal}
S.~Wei, Z.~Wu, and J.~Zhang, ``Multimodal motion control of magnetic continuum
  robot for endovascular intervention navigation,'' {\em IEEE/ASME Transactions
  on Mechatronics}, 2025.

\bibitem{linMagneticContinuumRobot2021}
D.~Lin, N.~Jiao, Z.~Wang, and L.~Liu, ``A magnetic continuum robot with
  multi-mode control using opposite-magnetized magnets,'' vol.~6, no.~2,
  pp.~2485--2492.

\bibitem{huang2024design}
Y.~Huang, Q.~Zhao, J.~Hu, and H.~Liu, ``Design and modeling of a multi-dof
  magnetic continuum robot with diverse deformation modes,'' {\em IEEE Robotics
  and Automation Letters}, vol.~9, no.~4, pp.~3956--3963, 2024.

\bibitem{wang2023magnetic}
Z.~Wang, D.~Weng, Z.~Li, L.~Chen, Y.~Ma, and J.~Wang, ``A magnetic-controlled
  flexible continuum robot with different deformation modes for vascular
  interventional navigation surgery,'' in {\em Actuators}, vol.~12, p.~247,
  MDPI, 2023.

\bibitem{cao2023magnetic}
Y.~Cao, Z.~Yang, B.~Hao, X.~Wang, M.~Cai, Z.~Qi, B.~Sun, Q.~Wang, and L.~Zhang,
  ``Magnetic continuum robot with intraoperative magnetic moment programming,''
  {\em Soft Robotics}, vol.~10, no.~6, pp.~1209--1223, 2023.

\bibitem{howell2001compliant}
L.~Howell, ``Compliant mechanisms. john wley \& sons,'' {\em Inc, New York},
  2001.

\bibitem{roesthuisSteeringMultisegmentContinuum2016}
R.~J. Roesthuis and S.~Misra, ``Steering of multisegment continuum manipulators
  using rigid-link modeling and fbg-based shape sensing,'' vol.~32, no.~2,
  pp.~372--382.

\bibitem{wu2025unified}
Z.~Wu, J.~Luo, S.~Wei, and J.~Zhang, ``Unified modeling and structural
  optimization of multi-magnet embedded soft continuum robots for enhanced
  kinematic performances,'' {\em arXiv preprint arXiv:2507.10950}, 2025.

\bibitem{kato2014tendon}
T.~Kato, I.~Okumura, S.-E. Song, A.~J. Golby, and N.~Hata, ``Tendon-driven
  continuum robot for endoscopic surgery: Preclinical development and
  validation of a tension propagation model,'' {\em IEEE/ASME Transactions on
  Mechatronics}, vol.~20, no.~5, pp.~2252--2263, 2014.

\bibitem{ikuta2006multi}
K.~Ikuta, H.~Ichikawa, K.~Suzuki, and D.~Yajima, ``Multi-degree of freedom
  hydraulic pressure driven safety active catheter,'' in {\em Proceedings 2006
  IEEE International Conference on Robotics and Automation, 2006. ICRA 2006.},
  pp.~4161--4166, IEEE, 2006.

\bibitem{kimTeleroboticNeurovascularInterventions2022}
Y.~Kim, E.~Genevriere, P.~Harker, J.~Choe, M.~Balicki, R.~W. Regenhardt, J.~E.
  Vranic, A.~A. Dmytriw, A.~B. Patel, and X.~Zhao, ``Telerobotic neurovascular
  interventions with magnetic manipulation,'' {\em Science Robotics}, vol.~7,
  p.~eabg9907, Apr. 2022.

\bibitem{milnor1963morse}
J.~W. Milnor, {\em Morse theory}.
\newblock No.~51, Princeton university press, 1963.

\bibitem{audin2014morse}
M.~Audin, M.~Damian, and R.~Ern{\'e}, {\em Morse theory and Floer homology},
  vol.~2.
\newblock Springer, 2014.

\bibitem{edelsbrunner2010computational}
H.~Edelsbrunner and J.~Harer, {\em Computational topology: an introduction}.
\newblock American Mathematical Soc., 2010.

\end{thebibliography}


\end{document}